\documentclass[conference]{IEEEtran}

\ifCLASSINFOpdf
   \usepackage[pdftex]{graphicx}
\else
   \usepackage[dvips]{graphicx}
\fi

\usepackage[cmex10]{amsmath}
\usepackage{amssymb}
\usepackage{amsthm}
\usepackage{mathtools}
\usepackage[hidelinks]{hyperref}
\usepackage{cite}
\usepackage{soul}
\setstcolor{red}
\usepackage{lipsum}
\usepackage{xargs}
\usepackage{booktabs}
\usepackage{multirow}
\usepackage{algorithm}
\usepackage{algpseudocode}
\usepackage{tabularx}
\usepackage[normalem]{ulem}
\useunder{\uline}{\ul}{}
\usepackage[table,xcdraw]{xcolor}
\usepackage{cite}
\usepackage[caption=false,font=footnotesize]{subfig}
\usepackage[textsize=tiny]{todonotes}
\usepackage{pgf-pie}
\usepackage{pgfplots}
\pgfplotsset{compat=1.18}
\usepgfplotslibrary{groupplots}

\makeatletter
\newcommand{\linebreakand}{%
  \end{@IEEEauthorhalign}
  \hfill\mbox{}\par
  \mbox{}\hfill\begin{@IEEEauthorhalign}
}
\makeatother

\begin{document}

\title{Data Enrichment Opportunities for Distribution Grid Cable Networks using Variational Autoencoders}



\author{

}

\author{\IEEEauthorblockN{Konrad Sundsgaard}
\IEEEauthorblockA{\textit{Dept. of Wind and Energy Systems} \\
\textit{Technical University of Denmark}\\
Copenhagen, Denmark\\
kosun@dtu.dk}
\and
\and
\IEEEauthorblockN{Kutay Bölat}
\IEEEauthorblockA{\textit{Dept. of Electrical Sustainable Energy} \\
\textit{Delft University of Technology}\\
Delft, The Netherlands\\
k.bolat@tudelft.nl
}
\and
\IEEEauthorblockN{Guangya Yang }
\IEEEauthorblockA{\textit{Dept. of Wind and Energy Systems} \\
\textit{Technical University of Denmark}\\
Copenhagen, Denmark \\
gyy@dtu.dk}
}

\maketitle
\global\csname @topnum\endcsname 0
\global\csname @botnum\endcsname 0
\begin{abstract}

Electricity distribution cable networks suffer from incomplete and unbalanced data, hindering the effectiveness of machine learning models for predictive maintenance and reliability evaluation. Features such as the installation date of the cables are frequently missing. To address data scarcity, this study investigates the application of Variational Autoencoders (VAEs) for data enrichment, synthetic data generation, imbalanced data handling, and outlier detection. 
Based on a proof-of-concept case study for Denmark, targeting the imputation of missing age information in cable network asset registers, the analysis underlines the potential of generative models to support data-driven maintenance. However, the study also highlights several areas for improvement, including enhanced feature importance analysis, incorporating network characteristics and external features, and handling biases in missing data. Future initiatives should expand the application of VAEs by incorporating semi-supervised learning, advanced sampling techniques, and additional distribution grid elements, including low-voltage networks, into the analysis. 


\end{abstract}

\section{Introduction}

The rising need for investment and modernization in power distribution systems requires robust tools to support decision-making and enhance maintenance strategies. Reliability models are one promising avenue for providing better insights into the condition of distribution grid components and supporting decision-making. Machine learning (ML) techniques offer high potential for transforming such reliability modelling of distribution grids\cite{Alliander2016}. They can model complex, non-linear relationships by leveraging historical failures and ageing drivers, eventually providing better prediction performance \cite{Merigeault2021}. 

In Denmark, medium-voltage (MV) cable networks are a highly applicable subject for such data-driven reliability studies due to their unique characteristics. These networks usually comprise older components, are less accessible than their high-voltage counterparts, and significantly contribute to grid outages. Previous research has predominantly concentrated on high-voltage assets and sensor-based solutions, underscoring the need for a targeted exploration of distributed, underground assets like MV cables \cite{MORTENSEN2024_Literature}. In Denmark, the two primary cable technologies—PILC (paper-insulated lead-covered) and XLPE (cross-linked polyethylene)—exhibit significantly different failure profiles. PILC cables show a steady increase in failure rates, driven by age-related degradation, while XLPE cables maintain relatively stable performance \cite{Hansen2023}. This disparity highlights the pressing need for forward-looking maintenance strategies beyond corrective measures.

Despite its promise, ML faces notable challenges. For example, data limitations significantly hinder appropriate modeling \cite{Sundsgaard2023}. Available data often lack critical features such as cable installation age, affected cable systems, and geographical information \cite{Sundsgaard2023_explore_data_collection}. The distributed nature of datasets complicates the combination of asset information, failure records, and location-based features, resulting in overfitting risks and reduced generalizability during conventional ML modeling \cite{Sundsgaard2023_ER_Model}.

Data enrichment techniques present a possible solution to these challenges. Furthermore, they might be particularly relevant for data representing underground MV cables. These components are inaccessible for visual inspection, a significant challenge to retrospective data collection. Established techniques like k-nearest neighbors (KNN), Iterative Imputer (IIm), and MissForest (MF) have been employed previously to address missing data. Recently, studies have highlighted the relevance of text-based classification and Large Language Models (LLMs) for missing data imputation. However, to the best of the author's knowledge, no specified studies have applied such techniques to medium voltage cables and failure data. In \cite{Sundsgaard2023_Text_AI}, the authors leverage text-based models to extract information from the textual information linked to failure registers. 
However, the applied techniques have several shortcomings. For example, they fail to tackle broader issues such as data imbalance and synthetic dataset creation, which are both critical for fostering data sharing and enhancing ML model development. Moreover, targeting the imputation to specific framework conditions is challenging due to the large volume of data required for model training. 

This paper proposes using variational autoencoders (VAEs) to address these challenges. VAEs, a type of generative model, excel at handling missing data and enabling the creation of balanced, synthetic datasets. This capability is particularly relevant for reliability studies of MV cable systems, where diverse and high-quality data are essential. By leveraging VAEs, this paper aims to explore new opportunities in predictive modeling and predictive maintenance, offering a scalable and data-driven approach to supporting the evolving needs of DSOs in Denmark.
\section{Issues of Missing Data in Reliability Modeling of Distribution Grid Cable Networks}
Failure data is crucial for making accurate and informed asset management and maintenance decisions regarding distribution grid components. For instance, the data is used to develop reliability models that predict failures in MV cables \cite{Faivre2019}. Unfortunately, DSOs often face data scarcity or lack of information about past component failures, as the following sections outline further. 

\subsection{Failure Reports}
Since 1974, Danish DSOs have recorded MV cable failures in the national fault and outage statistic (ELFAS) \cite{Sundsgaard2023}. Initially, the primary motivation for collecting the data was due to regulating requirements, such as documenting the continuity of supply. Consequently, to ensure the calculation of system reliability indices such as SAIDI or SAIFI, the failure reports strictly require parameters such as the number of interrupted customers, interruption duration, and failure dates. Many further parameters are also mandatory, such as failure reason or failure type, and thus also show high quality and availability. However, nowadays, the demand for enhanced condition assessment data is growing. 

For instance, one highly relevant feature for reliability assessment is the installation date of the failed cable, as this enables the calculation of the cable's age during the failure. Moreover, once available, features, such as the geographical details of the failure and information about the cable system involved, will significantly improve the reliability prediction of MV cables. Both enable the combination of failure reports with further datasets, such as asset registers, geographical aging drivers, or cable placement conditions. Unfortunately, registration of these parameters is currently optional, leading to shares of missingness, as Fig. \ref{fig:combined_piecharts_missingness} underlines.

\begin{figure}[htbp]
\centering
\begin{tikzpicture}
    \begin{axis}[
        ybar stacked,
        symbolic x coords={Affected Line, Installation Date, Failure Location},
        xtick=data,
        ymin=0,
        ymax=100,
        ylabel={Percentage (\%)},
        bar width=15pt,
        width=\linewidth,
        height=7cm,
        enlarge x limits=0.2,
        legend style={at={(0.5,-0.15)}, anchor=north, legend columns=-1},
        nodes near coords,
        nodes near coords align={vertical},
    ]
        \addplot[fill=green!70!black] coordinates {(Affected Line, 83) (Installation Date, 51) (Failure Location, 14)};
        \addplot[fill=red!70!black] coordinates {(Affected Line, 17) (Installation Date, 49) (Failure Location, 86)};
        \legend{Available, Not Available}
    \end{axis}
\end{tikzpicture}
\caption{Data availability for selected features in failure reports, sorted by availability.}
\label{fig:stacked_bar_chart_sorted}
\end{figure}
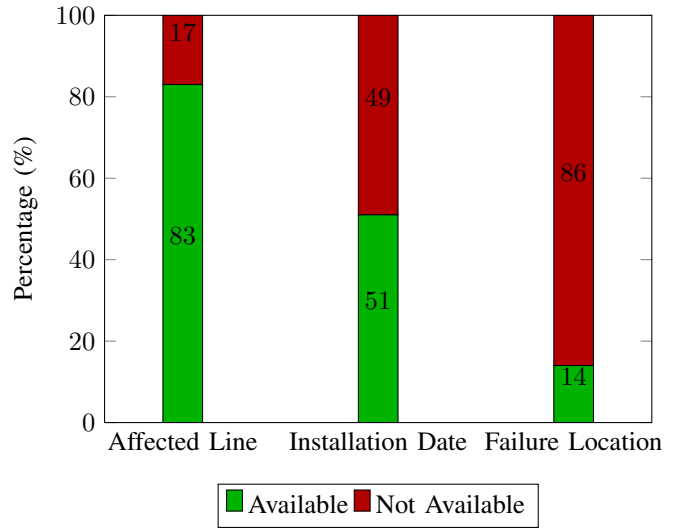





\subsection{Asset Information}
As shown in \cite{Sundsgaard2023_explore_data_collection}, one possibility to increase the available information on failure reports is to combine it with further datasets, such as asset registers managed by Danish DSOs. However, these datasets also contain missing information. Even though the level of data availability differed from different DSO environments and previous collection initiatives, information on the cable's installation age is also frequently missing. 

\section{Data Enrichment Opportunities}

The severe unavailability of different features in the ELFAS dataset limits its usability for downstream applications such as predictive reliability analysis. One vector to approach this problem is enriching (or balancing) the dataset by (1) modelling the dataset probabilistically and (2) taking adequate samples from the model. We propose to employ VAE-based models for this task thanks to their versatility in modelling complex probability distributions.

\subsection{Variational Autoencoders}
VAEs\cite{kingma2013auto,rezende2014stochastic} are deep learning-based generative models that employ deep neural networks to model complex data distributions. They consist of two main parts: encoder and decoder neural networks. The encoder maps the high dimensional data to a (preferably) low dimensional latent space while the decoder tries to reverse this process. However, unlike the conventional autoencoder models, VAEs perform this mapping probabilistically, an integral part of its renowned generative powers.

\subsection{Conditional Variational Autoencoders}
Their versatile neural network-based compositions allow VAEs' distribution modelling capabilities to be extended to conditional probability distributions \cite{sohn2015learning}. Such conditioning creates a controlled generation (sampling) mechanism when the practitioners are endowed with extra (observed) information or create guided scenarios. For example, a pool of asset data from different DSOs can be modelled as a conditional probability distribution where the condition can be the DSO identity. By this, the practitioner can generate data for a given DSO if they believe that the asset data's statistical characteristics depend on the DSO to which they belong. Note that the conditioning is not limited to single variable types but combinations of them, creating a trade-off between further controllability and data availability for training.

\begin{figure*}[!t]
    \centering
    \includegraphics[width=0.8\linewidth]{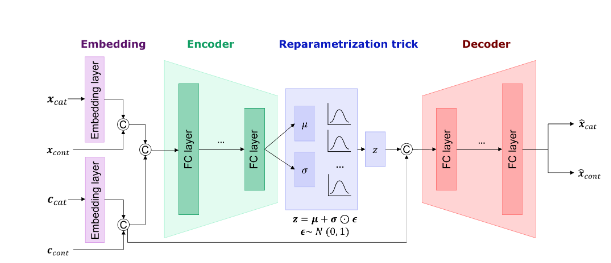}
    \caption{Computational diagram of a conditional CVAE. Embedding layers are applied to categorical inputs and conditions. \cite{Parra2023}}
    \label{fig:CVAE}
\end{figure*}

An overall diagram of a conditional VAE (CVAE) is illustrated in Fig. \ref{fig:CVAE}. As can be seen, the only modification on top of a regular VAE structure is concatenating the relative conditions to the data points (and latent vectors) before inputting them into the encoder (and the decoder). Another aspect to be inspected is the occurrence of the embedding layers in Fig. \ref{fig:CVAE}. These layers consist of learnable dictionaries where each element is mapped to a continuous vector. This is important for this study since the asset data we are interested in contains categorical variables that should be transformed into continuous variables to input into neural networks. 

\subsection{Imputation by Pseudo-Gibbs Sampling}
Besides their data generation and encoding capabilities, (C)VAEs stand out with their explicit representation of the probability distribution that they model. This feature creates further opportunities to use them not only for data generation but also for various probability distribution-based applications such as missing value imputation and anomaly detection. In our case study, missing value imputation plays a big part since there is a severe unavailability of installation dates of the cables in the records. These values can be sampled by manipulating the marginal data distribution, which is the distribution of the (C)VAE models.

Unfortunately, (C)VAEs can supply the marginal probability distribution only in the approximate form, not as an analytical expression. This limits the applicability of the distribution-based applications using the trained (C)VAE. However, thanks to its probabilistic foundations, both in inference and training, (C)VAEs can be used for missing data imputation in a theoretically found manner using pseudo-Gibbs sampling \cite{rezende2014stochastic, mattei2018leveraging}. This sampling mechanism enables the practitioner to input the data point with missing values into the model with an initial guess and refine it through iterative inferences. 

\subsection{Regression by Semi-Supervised Learning}

One drawback of the missing value imputation approach is that the (C)VAE must be trained with the complete data. In cases of severe missingness, most of the dataset is discarded, which means that the model fails to capture intrinsic dependencies among features due to the lack of representative data points. To circumvent this, the practitioner can follow a semi-supervised learning strategy in cases where most of the missingness results from one feature (in our case, the installation date). Figure \ref{fig:Loss_convergence} exemplary visualizes the data loss compared to supervised and seme-supervised training, strongly underlying the potential. 

There are various ways to adapt semi-supervised learning to (C)VAEs \cite{kingma2014semi}. All of them are based on separating the feature that creates the abundance of the missingness as a target variable and training a model that models the distribution of the remaining variables and the mapping from them to the target variables. This means the model learns the mapping whenever the target variable is available, while distribution modelling does not require such a condition and keeps learning even if the target variable is missing.

One might question why semi-supervised learning is followed when most missingness comes from a single feature, translating into a regression problem. The semi-supervised approach has two main benefits. Firstly, if there is more missingness in different features but not as severe as the dominant feature, assessing all of them as target variables becomes infeasible in terms of mapping (the target variable size can get larger than the regressor variable size). In the semi-supervised setting, this minor missingness can be discarded without affecting the unsupervised learning severely and can be imputed later using pseudo-Gibbs sampling. Secondly, (C)VAEs can work as nonlinear dimensionality reduction models thanks to their encoding capabilities and be used as feature extractors for more efficient supervised learning tasks. Thus, the supervised mapping part of the semi-supervised learning can be relatively simple with respect to an end-to-end deep learning regression model.

\section{Case Study Implementation}

To highlight (C)VAE's potential for data enrichment and synthetic data creation in the context of data-driven reliability modeling, this paper implements a "proof-of-concept" case study implementing a VAE that models asset information for MV cable sections for the Danish Case. As a starting point, the data is limited to the most relevant information. The features include parameters, such as the cable section length, the installation age of the cable, the voltage, operation distribution system operator, insulation material, conductor material, conductor size, and number of conductors, as Table \ref{tab:features} describes further. 

\begin{table}[!ht]
\small
\centering
\caption{\textit{Asset Data Features used for VAE development}}
\label{tab:ELFASmetadata}
\begin{tabular}{lll}
\toprule
\textbf{Column} & \textbf{Type} & \textbf{Data Origin} \\ \hline
Length [m] & float & DSOs \\  
Age[years] & float & DSOs \\  
Operation Voltage[kV] & category & DSOs \\  
DSO[name] & category & DSOs \\  
Insulation[name] & category & DSOs\\
Conductor Material & category & DSOs\\
Conductor Size & category & DSOs \\
Number of Conductors & category & DSOs \\

\bottomrule
\end{tabular}\label{tab:features}
\end{table}
\subsection{Implementation Set-up}
The VAE's modeling and training are done using \texttt{PyTorch}. In addition, the data preprocessing employs scalers and encoders from \texttt{sklearn} to standardize numerical features and encode categorical variables. Model performance and data imputation quality are benchmarked against state-of-the-art imputation techniques, such as \texttt{MissForest} and \texttt{KNNImputer}. Additionally, \texttt{MLflow} is employed for tracking experiments, recording hyperparameters, and storing results, enabling efficient model comparison and reproducibility. 

\subsection{VAE Loss Function}
The objective function for the (C)VAE, \(\mathcal{L}_{\text{VAE}}\), is formulated in \eqref{Lossfunction}, as a weighted combination of three components. The first component, \(\mathcal{L}_{\text{cont}}\), represents the loss associated with the continuous variables and is weighted by the parameter \(\alpha\). The second component, \(\mathcal{L}_{\text{cat}}\), accounts for the loss of the categorical variables and is weighted by \((1 - \alpha)\). The third term, \(\beta \cdot \mathcal{L}_{\text{KL}}\), adds the Kullback-Leibler (KL) divergence regularization factor and ensures that the learned latent distribution remains close to the prior distribution. In this configuration, \(\alpha \in [0, 1]\) is a hyperparameter that balances the contributions of the continuous and categorical losses while \(\beta\) scales the KL divergence.

\begin{equation}\label{Lossfunction}
\mathcal{L}_{\text{VAE}} = \alpha \cdot \mathcal{L}_{\text{cont}} + (1 - \alpha) \cdot \mathcal{L}_{\text{cat}} + \beta \cdot \mathcal{L}_{\text{KL}}
\end{equation}

The loss term \(\mathcal{L}_{\text{cont}}\), as \eqref{Losscontinious} further explains, represents the negative log-likelihood of the continuous variables under a Gaussian distribution with unit variance, averaged over all \(N\) samples. It penalizes the squared differences between the true values \(x_{i,j}\) and their reconstructions \(\hat{x}_{i,j}\).

\begin{equation}\label{Losscontinious}
\mathcal{L}_{\text{cont}} = -\frac{1}{N} \sum_{i=1}^N \sum_{j=1}^{d_{\text{cont}}} \left[ -\frac{1}{2} \log(2\pi) - \frac{1}{2}(x_{i,j} - \hat{x}_{i,j})^2 \right]
\end{equation}

Furthermore, \eqref{Losscategorical} shows that \(\mathcal{L}_{\text{cat}}\) equals the sum of the cross-entropy losses computed for each of the \(d_{\text{cat}}\) categorical variables. Consequently, it measures the discrepancy between the true categorical values \(x_{\text{cat},k}\) and their predicted probabilities \(\hat{x}_{\text{cat},k}\).

\begin{equation}\label{Losscategorical}
\mathcal{L}_{\text{cat}} = \sum_{k=1}^{d_{\text{cat}}} \text{CrossEntropy}(\hat{x}_{\text{cat},k}, x_{\text{cat},k})
\end{equation}
%

\subsection{VAE - Asset Data }
\subsubsection{Model Training}
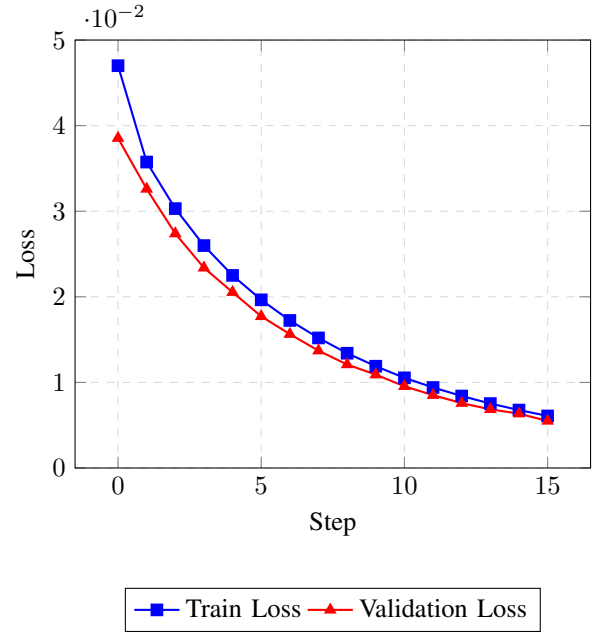
\begin{figure}[!ht]
    \centering
\begin{tikzpicture}
    \begin{axis}[
        xlabel={Step},
        ylabel={Loss},
        legend style={at={(0.5,-0.28)},anchor=north,legend columns=-1},
        grid=major,
        grid style={dashed,gray!30},
        ymin=0,
        ymax=0.05
    ]

    \addplot[
        color=blue,
        thick,
        mark=square*
    ]
    coordinates {
        (0,0.04700355720524164)
        (1,0.03574577334982695)
        (2,0.030308938538760863)
        (3,0.025984695001558575)
        (4,0.02249834524503526)
        (5,0.019647918847484545)
        (6,0.017236217296933705)
        (7,0.015193206970896407)
        (8,0.013405621744654768)
        (9,0.011889552126733942)
        (10,0.010534016025982255)
        (11,0.009406285623032466)
        (12,0.008409005085188638)
        (13,0.007527661869681526)
        (14,0.006770168520949228)
        (15,0.006080087705573766)
    };
    \addlegendentry{Train Loss}

    \addplot[
        color=red,
        thick,
        mark=triangle*
    ]
    coordinates {
        (0,0.038539594587195844)
        (1,0.03260185446789133)
        (2,0.02738354769989227)
        (3,0.02338793683548315)
        (4,0.02053711384603472)
        (5,0.01771530482188118)
        (6,0.01563550704037232)
        (7,0.013705097758225562)
        (8,0.012090743300304862)
        (9,0.010903858451609537)
        (10,0.009547367320970146)
        (11,0.008506465848129472)
        (12,0.007576400780519285)
        (13,0.0068513375370224715)
        (14,0.006345403402666896)
        (15,0.00549365041029374)
    };
    \addlegendentry{Validation Loss}

    \end{axis}
\end{tikzpicture}
    \caption{Training and Validation Loss of the VAE (asset data) over different training epochs. }
    \label{fig:Loss_convergence}
\end{figure}

Figure \ref{fig:Loss_convergence} illustrates the reduction in loss across the training epochs. Given the extended training duration and the primary focus on demonstrating a proof of concept, the hyperparameter space was constrained during optimization, and model training was terminated prior to achieving complete convergence. The hyperparameters employed for this training run are detailed in Table \ref{tab:parameters}. 

\begin{table}[!ht]
\caption{Selected hyperparameters for VAE (asset data) training.}
\centering
\begin{tabular}{|l|l|}
\hline
\textbf{Parameter}    & \textbf{Value} \\ \hline
alpha                 & 0.07127        \\ \hline
batch\_size           & 128            \\ \hline
beta                  & 0.0275         \\ \hline
decoder\_params       & 74651          \\ \hline
embedding\_params     & 167            \\ \hline
encoder\_params       & 63815          \\ \hline
hidden\_dim           & 145            \\ \hline
latent\_dim           & 13             \\ \hline
learning\_rate        & 0.0001         \\ \hline
optimiser             & Adam           \\ \hline
\end{tabular}

\label{tab:parameters}
\end{table}

\subsubsection{Model Validation}
\begin{figure}[!htb]
    \centering
    \includegraphics[width=0.8\linewidth]{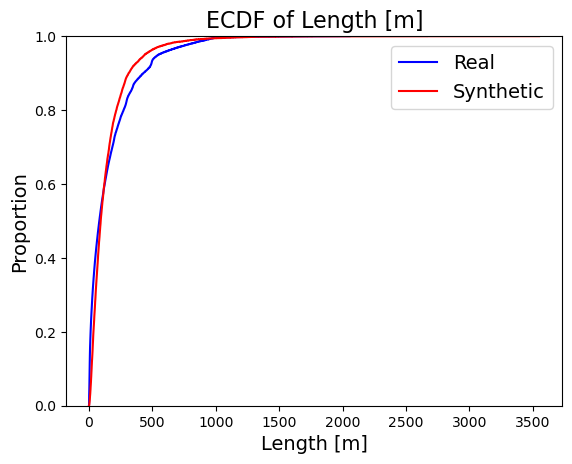}
    \caption{Empirical Cumulative Distribution Function (ECDF) for the continuous feature: cable section length. Computed during sample level testing}
    \label{fig:ecdf_length}
\end{figure}

\begin{figure}[!htb]
    \centering
    \includegraphics[width=0.8\linewidth]{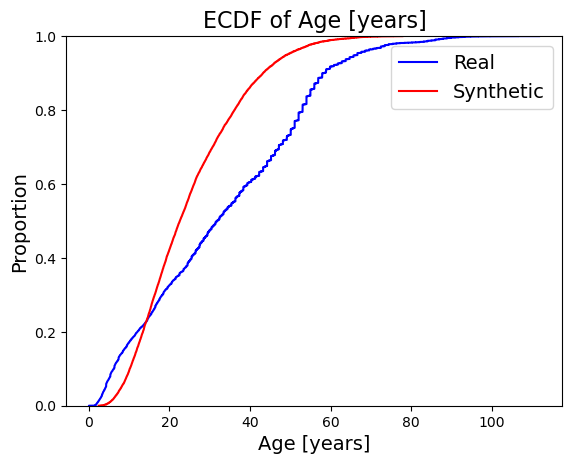}
    \caption{Empirical Cumulative Distribution Function (ECDF) for the continuous feature: installation age. Computed during sample level testing}
    \label{fig:ecdf_age}
\end{figure}

\begin{figure}[!htb]
    \centering
    \includegraphics[width=0.8\linewidth]{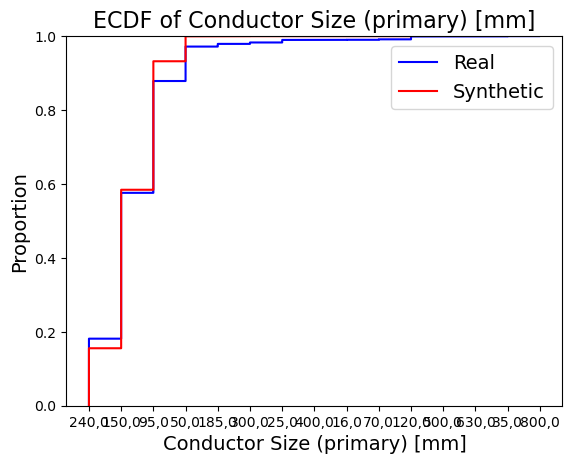}
    \caption{Empirical Cumulative Distribution Function (ECDF) for the categorical feature: conductor size. Computed during sample level testing}
    \label{fig:ecdf_conductor}
\end{figure}

The primary objective of (C)VAE models is to generate samples that mimic the joint probability distribution of the original data. Consequently, the next step in the pipeline involves sampling from the trained distributions, plotting a histogram or density plot, and comparing the results with the original data distribution. The simplest way to compare the distributions is by visually analyzing their shape, spread, and central tendencies. Figures \ref{fig:ecdf_length}, \ref{fig:ecdf_age}, and \ref{fig:ecdf_conductor} provide a visualization of the Empirical Cumulative Distribution Function (ECDF) of the continuous features' installation age, length, and the categorical feature conductor material. The visualizations underline graphically that the main characteristics are captured during training without replicating the distributions.  

Similarly, Table \ref{tab:real_vs_synthetic_compact} compares the actual and synthetic distribution of the continuous features' age and length based on statistics. The table highlights real and synthetic data similarity, showing relatively close mean and standard deviation values and hardly any differences in statistical distance metrics like the Kolmogorov-Smirnov (KS) statistic \cite{Smirnov}.

\begin{table}[!ht]
\centering
\caption{Real vs. Synthetic Data (Mean ± Std Dev, KS Stat)}
\begin{tabular*}{\linewidth}{@{\extracolsep{\fill}}lccc}
\toprule
\textbf{Feature} & \textbf{Real Data} & \textbf{Synthetic Data} & \textbf{KS Stat} \\
\midrule
LogAge [years]   & 3.20 ± 0.90 & 3.20 ± 0.76 & 0.13 \\
LogLength [m]    & 4.11 ± 1.64 & 4.14 ± 1.43 & 0.06 \\
Age [years]      & 33.12 ± 20.87 & 30.40 ± 16.29 & 0.13 \\
Length [m]       & 158.99 ± 202.79 & 131.64 ± 147.34 & 0.06 \\
\bottomrule
\end{tabular*}
\label{tab:real_vs_synthetic_compact}
\end{table}

\subsubsection{Amputation Testing}
One common issue in imputation tasks is that the ground truth is unknown, making it challenging to evaluate imputation performance. Therefore, this paper follows the strategy of testing the VAE's imputation capability by amputating. In this context, amputation refers to systematically removing a substation portion of the data while maintaining its underlying structure. This mimics realistic missingness in the data, with the difference that original data can be used to validate the imputation performance.

The core motivation of VAE training is identifying missing installation ages of the cable section. As elaborated, the installation age is the most frequently missing feature. Consequently, for the asset data imputation, the installation data is systematically removed from the validation set and imputed again with the help of the VAE.

During amputation testing, the VAE's imputation capability is benchmarked against simpler imputation techniques. This allows for determining whether the VAE's imputation performance is superior and, if so, evaluating whether it justifies the VAE training efforts. To achieve this, widely accepted statistical performance metrics, including the Mean Absolute Error (MAE), the Root Mean Square Error (RMSE), and the R-squared Score, are employed. 

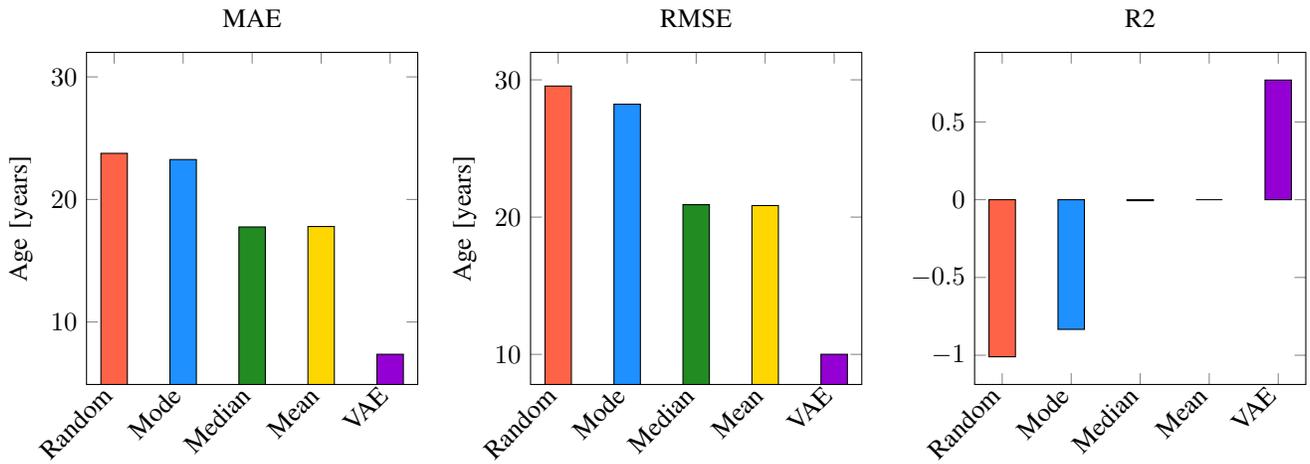
\begin{figure*}[!ht]
    \centering
    
    \begin{tikzpicture}
        \begin{groupplot}[
            group style={
                group size=3 by 1,
                horizontal sep=1.5cm,
                vertical sep=1.5cm,
            },
            height=6cm,
            width=0.33\linewidth,
            ylabel style={align=center},
            xlabel style={align=center},
            xticklabel style={rotate=45, anchor=east},
            legend style={at={(1.05,1)}, anchor=north west, draw=none},
        ]

        \definecolor{randomcolor}{RGB}{255,99,71} 
        \definecolor{modecolor}{RGB}{30,144,255}  
        \definecolor{mediancolor}{RGB}{34,139,34} 
        \definecolor{meancolor}{RGB}{255,215,0}  
        \definecolor{vaecolor}{RGB}{148,0,211}   

        \nextgroupplot[
            title=MAE,
            ylabel={Age [years]},
            ymax=32,
            xtick={1,2,3,4,5},
            xticklabels={Random,Mode,Median,Mean,VAE},
        ]
        \addplot[ybar,fill=randomcolor] coordinates {(1,23.761665)};
        \addplot[ybar,fill=modecolor] coordinates {(2,23.254992)};
        \addplot[ybar,fill=mediancolor] coordinates {(3,17.757384)};
        \addplot[ybar,fill=meancolor] coordinates {(4,17.792777)};
        \addplot[ybar,fill=vaecolor] coordinates {(5,7.367093)};


        \nextgroupplot[
            title=RMSE,
            ylabel={Age [years]},
            ymax=32,
            xtick={1,2,3,4,5},
            xticklabels={Random,Mode,Median,Mean,VAE},
        ]
        \addplot[ybar,fill=randomcolor] coordinates {(1,29.548152)};
        \addplot[ybar,fill=modecolor] coordinates {(2,28.222190)};
        \addplot[ybar,fill=mediancolor] coordinates {(3,20.905618)};
        \addplot[ybar,fill=meancolor] coordinates {(4,20.841326)};
        \addplot[ybar,fill=vaecolor] coordinates {(5,10.005951)};

        \nextgroupplot[
            title=R2,
            xtick={1,2,3,4,5},
            xticklabels={Random,Mode,Median,Mean,VAE},
        ]
        \addplot[ybar,fill=randomcolor] coordinates {(1,-1.010064)};
        \addplot[ybar,fill=modecolor] coordinates {(2,-0.833710)};
        \addplot[ybar,fill=mediancolor] coordinates {(3,-0.006179)};
        \addplot[ybar,fill=meancolor] coordinates {(4,0.000000)};
        \addplot[ybar,fill=vaecolor] coordinates {(5,0.769503)};
        
        \end{groupplot}
    \end{tikzpicture}
    \caption{Benchmarking the metrics of VAE's imputation performance against simpler imputation methods such as random and informed random imputation. }
    \label{fig:benchmarking_imputation}
\end{figure*}

Figure \ref{fig:benchmarking_imputation} compares the imputation performance of the VAE during amputation. It is observed that the VAE strongly outperforms random imputation and informed random strategies, such as mode, median, and mean imputation. 

\begin{figure*}[!ht]
    \centering
    \begin{tikzpicture}
        \begin{groupplot}[
            group style={
                group size=3 by 1,
                horizontal sep=1.5cm,
                vertical sep=1.5cm,
            },
            height=6cm,
            width=0.33\linewidth,
            ylabel style={align=center},
            xlabel style={align=center},
            xticklabel style={rotate=45, anchor=east},
            legend style={at={(1.05,1)}, anchor=north west, draw=none},
        ]

        \definecolor{vaecolor}{RGB}{148,0,211}   
        \definecolor{knncolor}{RGB}{255,165,0}   
        \definecolor{itercolor}{RGB}{238,130,238} 
        \definecolor{mfcolor}{RGB}{165,42,42}    

        \nextgroupplot[
            title=MAE,
            ylabel=Value,
            ymin=0,
            ymax=15,
            xtick={1,2,3,4},
            xticklabels={KNN,IIm,MF,VAE},
        ]
        \addplot[ybar,fill=knncolor] coordinates {(1,7.4)};
        \addplot[ybar,fill=itercolor] coordinates {(2,7.6)};
        \addplot[ybar,fill=mfcolor] coordinates {(3,7.8)};
        \addplot[ybar,fill=vaecolor] coordinates {(4,7.367093)};


        \nextgroupplot[
            title=RMSE,
            ylabel=Value,
            ymin=0,
            ymax=15,
            xtick={1,2,3,4},
            xticklabels={KNN,IIm,MF,VAE},
        ]
        \addplot[ybar,fill=knncolor] coordinates {(1,9.7)};
        \addplot[ybar,fill=itercolor] coordinates {(2,10.2)};
        \addplot[ybar,fill=vaecolor] coordinates {(4,10.005951)};
        \addplot[ybar,fill=mfcolor] coordinates {(3,10.7)};

        \nextgroupplot[
            title=R2,
            ylabel=Value,
            ymin=0,
            xtick={1,2,3,4},
            xticklabels={KNN,IIm,MF,VAE},
        ]
        \addplot[ybar,fill=knncolor] coordinates {(1,0.62)};
        \addplot[ybar,fill=itercolor] coordinates {(2,0.71)};
        \addplot[ybar,fill=vaecolor] coordinates {(4,0.769503)};
        \addplot[ybar,fill=mfcolor] coordinates {(3,0.71)};
        
        \end{groupplot}
    \end{tikzpicture}
    \caption{Benchmarking the metrics of VAE's imputation performance against more sophisticated state-of-the-art imputation methods such as KNN, ITer, and MF.}
    \label{fig:imputation_metrics}
\end{figure*}
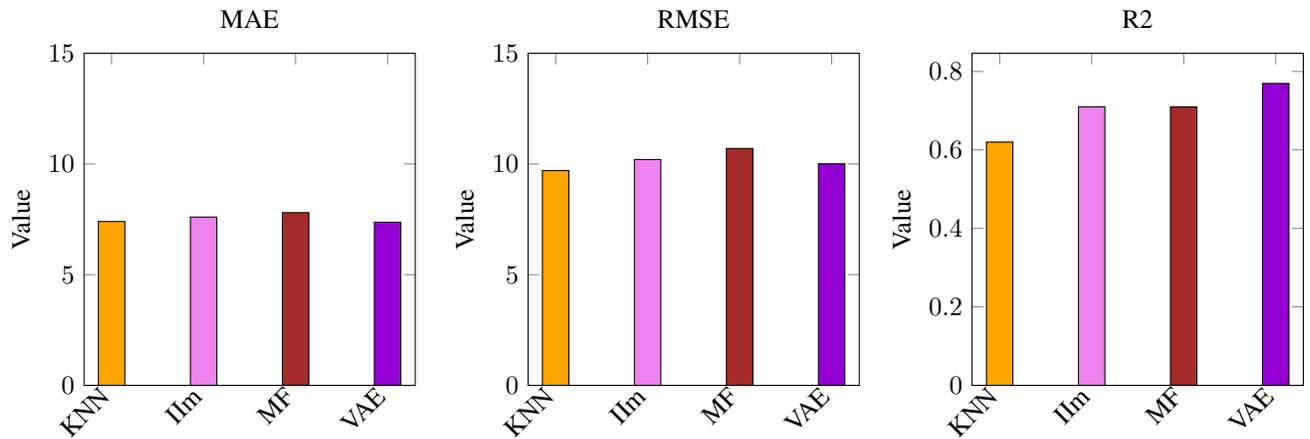

However, a fair comparison is only achieved if the results are benchmarked against more sophisticated imputation methods such as KNN, IIm, and MissForest. Figure \ref{fig:imputation_metrics} provides such a comparison and reveals that, even though the VAE ranks best among the approaches, it does not significantly outperform the other approaches. This indicates that further investigations and features relating to the installation age of the cables are required to improve imputation performance.

\section{Discussion \& Future Work}
This paper highlights the potential of generative models, such as (C)VAEs, for missing data imputation, handling of imbalanced data, and synthetic data generation in reliability studies of MV cables. However, the results presented also underscore the need for significant model improvements and further development. Several opportunities exist to advance the use of VAE-based data enrichment for distribution grid data and reliability studies, which are further discussed below. 

For example, future research should prioritize investigating the importance of the selected features linked to imputation performance, as certain features are likely to impact model predictions more than others. By systematically evaluating the contributions of existing and newly proposed features, models can be refined to achieve greater quality. Furthermore, assessing features' individual contributions might help DSOs decide what data to collect and deepen their understanding of underlying data relations. 

One core requirement for future work is to apply real imputation, not amputation, and investigate the impact of the presented data enrichment techniques on practical applications within the context of reliability modeling. By formulating proper downstream tasks, such as ranking MV cables after their risk of failures, the impact of generative data enrichment should be revealed, e.g., by training the model on the original and enriched dataset. In this regard, future studies should also investigate the possibilities of generative models, such as (C)VAEs, on how to cope with imbalanced data issues. 

One challenge in this regard is that the distribution of the training data may differ from that of the imputed data. Since missing data can arise from different types of missingness—Missing Completely at Random (MCAR), Missing at Random (MAR), or Missing Not at Random (MNAR)—it can be difficult to accurately determine the type present. In the context of age prediction, for instance, missingness is likely biased toward older cables, suggesting MNAR. Addressing such biases during training remains an open research question, particularly in determining whether the available training data is sufficient to detect and account for these dependencies. 

The installation age of the cable is not the only missing information. There are many other features that contain data scarcity issues. As a result, future work should expand the analysis and evaluation of the imputation performance to further features. Additionally, so far, the investigation has concentrated on either the asset information or failure reports. However, imputation might likely be improved using combined datasets. Consequently, future work must investigate how (C)VAEs could contribute to existing challenges in data combination and how the addition of further features, such as asset information from connected substations, information on cable placement conditions, weather information, and digging registers might benefit (C)VAE modeling. 

Eventually, (C)VAEs' ability to sample from trained probability distributions could facilitate the generation of a more balanced and larger data set, which might serve as a benchmark for downstream tasks across different DSOs, such as developing reliability models for MV cables. However, as some of the relevant data are highly sensitive, future work must also integrate solutions that maintain and ensure data privacy. 

This study chose to focus on (C)VAE during the study. However, other approaches exist, such as Generative Adversarial Networks, which could be applied to the problem in a similar manner. Consequently, future work should ensure a proper benchmark of the approaches to select the most appropriate technique depending on the modeling requirements. Similarly, future work could incorporate more sophisticated sampling methods within the process of missing value imputation, such as Advanced Coupled-Metropolis-within-Gibbs (AC-MWG), to name only one.   

This study focuses on MV cables as the component of interest. However, even though they represent a highly distributed and relevant component category for reliability studies, the amount and complexity of training data are limited compared to even further distributed components, such as low-voltage cables. 

Therefore, future work could apply similar techniques to low-voltage cable asset data and failure reports. Integrating the network characteristics in the (C)VAE model training provides another interesting reflection. For example, this could be done by transforming the training data into graph structures and applying network-based methods, such as Variational Graph Auto-Encoders, to the problem. In this way, network dependencies, such as similar installation times for connected grid areas, might be revealed, further improving imputation performance. 

Finally, future efforts must recognize the capability of (C)VAEs in detecting outliers. Implementing (C)VAE-driven data cleaning processes could enhance data quality and improve model performance. This data cleaning is especially important for low-voltage assets, where managing a large number of components makes effective data collection and management particularly challenging.

\section{Conclusion}
This research introduces Variational Autoencoders (VAEs) as a suitable approach to addressing data issues in the reliability modeling of medium-voltage (MV) cable networks. 
After outlining VAEs' data enrichment possibilities, such as missing value imputation and class balance, and integrating them in the data pipeline for power grid failure and asset data, the work narrows down to the application of VAE and (C)VAEs for missing value imputation. 

Accordingly, the study establishes a Proof of Concept for employing VAEs to impute missing information about the cable's installation age. 

A visual and statistical comparison reveals that the VAE model successfully replicates the original data distributions. Furthermore, the amputation capabilities of the trained models are promising. The VAE outperformed simple imputation techniques during benchmarking, such as random and informed random imputation. Also, compared to more sophisticated state-of-the-art imputation methods, the VAE demonstrates competitive performance. 

Consequently, the proof-of-concept implementation highlights VAEs' capability to perform targeted imputations and create enriched datasets. However, the study also identifies several limitations and areas for improvement, such as enhancing feature selection, addressing data biases, and integrating network-specific characteristics.

Future research should combine datasets to capture more comprehensive asset information, incorporate advanced sampling methods, and extend the work to more distribution grid components, such as low-voltage cable networks. By addressing these opportunities, VAEs have the potential to transform data-driven maintenance strategies, ultimately leading to improved management decisions and budget allocation.

\section{Acknowledgments}
This work builds upon a master thesis fulfilled by Eric Planas I Parra at the Technical University of Denmark titled: "Data enrichment strategies for AI-based reliability assessment of distribution grid components" \cite{Parra2023}. 
This manuscript has been developed within the framework of the \href{https://innocypes.eu/}{InnoCyPES} project, financially supported by the European Union's Horizon 2020 research and innovation program, through the Marie Sklodowska Curie Grant agrmt. No 956433. 

\bibliographystyle{IEEEtran}
\bibliography{refs.bib}

\end{document}